\pdfoutput=1

\documentclass[11pt]{article}

\usepackage[final]{acl}
\usepackage{enumitem}
\usepackage{times}
\usepackage{latexsym}

\usepackage[T1]{fontenc}

\usepackage[utf8]{inputenc}

\usepackage{microtype}

\usepackage{inconsolata}

\usepackage{graphicx}

\usepackage{hyperref}
\usepackage{url}
\usepackage{tabularx}
\usepackage{subcaption}
\usepackage{booktabs}
\usepackage{multirow}
\usepackage{adjustbox}
\usepackage{wrapfig}
\usepackage{amsmath}
\usepackage{hyperref}
\usepackage{cleveref}
\usepackage{xfrac}
\usepackage{xcolor}
\usepackage{colortbl}

\crefformat{section}{\S#2#1#3}
\crefformat{subsection}{\S#2#1#3}
\crefformat{subsubsection}{\S#2#1#3}

\usepackage{xcolor}
\definecolor{blue}{HTML}{4285f4} 
\definecolor{darkblue}{HTML}{0b5394} 
\definecolor{lightblue}{HTML}{9fc5e8} 
\definecolor{lightblue2}{HTML}{d7eff8}
\definecolor{lightblue3}{HTML}{cfe2f3} 
\definecolor{lightcornflowerblue2}{HTML}{a4c2f4} 
\definecolor{lightcornflowerblue3}{HTML}{c9daf8} 
\definecolor{darkcornflowerblue3}{HTML}{1c4587} 

\definecolor{orange}{HTML}{ff9900} 
\definecolor{lightorange}{HTML}{f9cb9c} 
\definecolor{lightorange3}{HTML}{fce5cd} 
\definecolor{darkorange}{HTML}{FF8C00} 
\definecolor{darkorange1}{HTML}{e69138} 
\definecolor{lightyellow2}{HTML}{f5efc4}
\definecolor{lightyellow3}{HTML}{fff2cc}

\usepackage{tcolorbox}
\usepackage[textwidth=3cm,textsize=scriptsize]{todonotes}

\colorlet{tablerowcolor}{gray!10}
\newcommand{\rowcol}{\rowcolor{tablerowcolor}}

\newcommand*\samethanks[1][\value{footnote}]{\footnotemark[#1]}

\title{Exploring Design Choices for Building Language-Specific LLMs}

\author{Atula Tejaswi\thanks{\;Authors contributed equally}, Nilesh Gupta\samethanks, Eunsol Choi \\ Department of Computer Science \\ The University of Texas at Austin \\ \texttt{\{atutej, nileshgupta2797, eunsol\}@utexas.edu}}

\begin{document}
\maketitle
\begin{abstract}
Despite rapid progress in large language models (LLMs), their performance on a vast majority of languages remains unsatisfactory. In this paper, we study building language-specific LLMs by adapting monolingual and multilingual LLMs. We conduct systematic experiments on how design choices (base model selection, vocabulary extension, and continued pretraining) impact the adapted LLM, both in terms of efficiency (how many tokens are needed to encode the same amount of information) and end task performance. We find that (1) the initial performance of LLM does not always correlate with the final performance after the adaptation. Adapting an English-centric models can yield better results than adapting multilingual models despite their worse initial performance on low-resource languages. (2) Efficiency can easily improved with simple vocabulary extension and continued pretraining in most LLMs we study, and (3) The optimal adaptation method (choice of the base model, new vocabulary size, training data, initialization strategy) is highly language-dependent, and the simplest embedding initialization works well across various experimental settings. Together, our work lays foundations on efficiently building language-specific LLMs by adapting existing LLMs.

\end{abstract}

\section{Introduction}

The predominance of English data on the internet, combined with a financial interest in English-centric applications, has led to the development of primarily monolingual LLMs~\cite{touvron2023llama, jiang2023mistral, abdin2024phi3} which exhibit significantly higher proficiency in English compared to other languages. Even when LLMs support other languages, their performance lags behind -- both in terms of end task performance and efficiency, measured by the amount of tokens required to encode information~\citep{ahia2023all}. 

Prior work has mainly focused on building multilingual models that cover a broad spectrum of languages~\citep{workshop2023bloom,ustun2024aya,lin2021few}, or building language-specific LLMs from scratch~\citep{zeng2023glm130b, muller2022cedille, sengupta2023jais}. Training a new language-specific LLM from scratch can be expensive, both in terms of compute and data requirements.
Therefore, recent efforts have focused on adapting existing, high-performing LLMs~\citep{cui2023efficient, rakutengroup2024rakutenai7b}, which consist of a two-stage process: (1) Adapting a model's tokenizer with tokens from the target language~\citep{cui2023efficient} to improve efficiency, and (2) updating model parameters through continued pre-training (CPT) to improve end task performance. 

For this straightforward adaptation recipe (illustarted in ~\autoref{fig:intro}), many design choices can affect the final performance, providing trade-offs between efficiency and downstream performance. We focus on three major design choices -- the choice of base LLM, the size of augmented vocabulary, and the amount of continued pretraining data. We empirically evaluate how these design choices impact the final task performances on four diverse languages (Hindi, Turkish, Arabic and Tamil) on multilingual benchmarks containing up to 7 tasks. Our goal is to provide guidance on how to build language-specific LMs based on our experimental results spanning seven base LMs.\footnote{Code is available at: \url{https://github.com/atutej/token-language-adaptation}} 

\noindent We summarize our main findings here:

\begin{figure*}
     \centering
     \includegraphics[width=\textwidth]{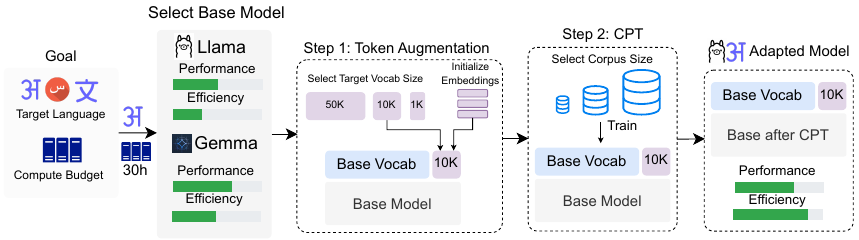}
     \caption{Building a language-specific language model. After selecting a base language model (LM), we adapt it with two main steps: 1) Token Augmentation, which primarily involves extending the tokenizer to a target vocabulary size, and 2) Continued Pre-Training on the target corpus.}
     \label{fig:intro}
\end{figure*}

\begin{itemize}[noitemsep,leftmargin=4mm]
    \item The base LM performance prior to adaptation is not always indicative of the final performance. Despite its limited initial performance, monolingual models (such as LLaMA-2) can be adapted effectively i.e. achieve comparable performance to base multilingual models after training on 200M target language tokens.
    \item A moderate amount of vocabulary addition (10K) is sufficient to close the gap between English and low-resource languages in terms of efficiency (how many tokens are required to encode the same amount of information). 
    \item Vocabulary extension initially drops the end task performance, but most base LMs (except the best multilingual model, Gemma-7B) can recover and improve end task performances after continued training on 200M target language tokens.\looseness=-1
    \item The initialization of new token parameters is important for efficient adaptation. Initializing new token embeddings with a mean of constituent token embeddings~\citep{liu2023enhancing}, is as good as more sophisticated initialization strategies~\cite{dobler-de-melo-2023-focus}.

    \item Despite some general patterns, adaptation performance is language and base LM specific, especially with respect to the pre-training corpus of base LMs. 
\end{itemize} 

Together, we lay foundation for studying language-specific adaptation of existing LLMs, enabling LLM access to a wider population. 

\section{Related Work}
\label{sec:related_work}

\paragraph{Token Adaptation}
Early work on adapting tokenizers focused on downstream tasks \citep{sachidananda-etal-2021-efficient} or domains \citep{hiraoka2020optimizing, hiraoka-etal-2019-stochastic} within a single language. \citet{liu2023enhancing} proposed to add new tokens trained from a specific domain (mental health QA) to enhance its generation speed. Similar to our work, \citet{dagan2024getting} provides a comprehensive study for optimizing tokenizers for LLMs built on code corpus. Our focus is on multilingual adaptation instead of a specific domain.

\paragraph{Cross-Lingual Transfer}
Recent efforts explore cross-lingual transfer of pre-trained English LLMs to new languages. \citet{husain2024romansetu} transliterate Indic languages to the Latin script to transfer linguistic capabilities. Recent work~\cite{zhao2024llama} showed that training on millions of target language tokens without vocabulary extension can match the end task performance of state-of-the art model trained on billions of tokens. However, this comes at the cost of inference efficiency.
Most similar to ours, a line of work~\cite{csaki2023efficiently, cui2023efficient,tikhomirov2023impact, lin2024mala500} perform vocabulary extension, showing it can improve generation efficiency on the target language. However, to the best of our knowledge, there is no unified perspective on the design choices such as the base model, vocabulary size when working with limited compute and data resources, and our study is the first that explores these aspects on multiple languages.\looseness=-1

Previous approaches have studied ways to optimally initialize embeddings for cross-lingual transfer. Some methods have only been applied to encoder-only models \citep{ebrahimi-kann-2021-adapt, minixhofer-etal-2022-wechsel}, others typically use external lexicons \citep{zeng2023greenplm, wang-etal-2022-expanding}, focus primarily on Latin scripts \citep{de-vries-nissim-2021-good}, or require training a secondary model/embeddings separately \citep{dobler-de-melo-2023-focus, ostendorff2023efficient}. We compare few initialization approaches, showing that the simple mean embedding initialization works well.
\begin{table*}
\small
\centering
\addtolength{\tabcolsep}{-2pt}
\begin{tabular}{lccccc}
\toprule
\textbf{Model} & \textbf{Vocab Size} & \textbf{Training Data} & \textbf{\# Training Tokens} & \textbf{\# Languages} & \textbf{\# Parameters} \\ \midrule \midrule
XGLM-7.5B & 256K & CC100-XL & 500B & 30 & 7.5B \\
Gemma-2B & 256K & unknown & 2T & unknown & 2.5B \\
Gemma-7B & 256K & unknown & 6T & unknown & 8.5B \\
Bloom-7.1B & 251K & ROOTS & 350B & 46 & 7.1B \\
Phi-2 & 50K & Synthetic + Web & 1.4T & unknown & 2.7B \\
Mistral-7B & 32K & unkown & unkown & unknown & 7.2B \\
LLaMA-2-7B & 32K & unknown & 2T & unknown & 6.7B \\ \midrule
\rowcol Adapted (Ours) & Base + 1K-50K & mC4 (subset, CPT) & Base + 100M-500M & Base + 1 & Base + 4M-200M \\
\bottomrule
\end{tabular}
\caption{The overview of LLMs compared in this work. For our adapted models, we add between 1K to 50K additional tokens into the vocabulary and continue training on $\sim$100M-500M tokens.}
\label{table:comparison_systems}
\end{table*}

\section{Method: Adapting LLM to a Target Language}

We introduce a straightforward adaptation process -- we will first generate language-specific tokens that will be added to the base vocabulary (\cref{subsec:augmenting_vocab}) of the model. Then, we continue training LMs with language modeling objective on the target language corpus such that they can make use of new tokens efficiently (\cref{subsec:continued_pretraining}).

\subsection{Augmenting Token Vocabulary}
\label{subsec:augmenting_vocab}

\paragraph{Generating Target Language Tokens} We train a BPE sentencepiece tokenizer \citep{kudo2018sentencepiece}\footnote{\url{https://github.com/google/sentencepiece}} using 300K examples (i.e. documents) from the mC4 corpus \citep{raffel2023exploring} on the target language, which yields a language specific vocabulary $V'$ with a target vocab size $|V'|$. We vary the target vocab sizes from 1K to 50K. 

\paragraph{Merging with Original Vocabulary} Let $\Delta V = V' - V$ denote the non-overlapping tokens from the new vocabulary with respect to the original vocabulary $V$. We append these tokens to the original as $V_{\text{new}} \leftarrow  V \oplus \Delta V$. Here, $\oplus$ denotes the concatenation operation, which implies that all new tokens are assigned lower priorities than those in the default vocabulary i.e. we assume that the frequency of the first ‘new’ token is lower than the last ‘old’ token in the BPE merging procedure~\citep{sennrich2016neural}. We preserve all the tokens from the original vocabulary, as opposed to discarding those with low-priority scores \citep{csaki2023efficiently}. We experimented with assigning higher priorities to the extra tokens, but found that it does not lead to significant gains. The resulting effective vocabulary size can be found in \autoref{table:stats} in the appendix.\looseness=-1

\subsection{Integrating New Tokens to the LLM}
\label{subsec:continued_pretraining}

\paragraph{Embedding Initialization} We initialize the token embeddings from the generated vocabulary $\Delta V$ as the mean embedding of its constituent tokens from the original tokenizer $V$, following prior work~\citep{liu2023enhancing, gee-etal-2022-fast}. Formally, the token embedding $E(v), \forall v \in \Delta V$ is obtained as,\looseness=-1
\begin{align}
    t = \text{Tokenize}(v\,; V); \, 
    E(v) =\frac{1}{|t|}\sum^{|t|}_{i}E(t_i)
\end{align}
where $E$ is the existing token embedding. Note, for models like LLaMA-2 which use separate un-embedding (LM head) parameters, we perform the same operation separately for the un-embedding layer. We also experiment warm starting the newly introduced parameters on a tiny fraction of the dataset ($10M$ tokens) through continued pre-training \citep{downey2023embedding}, keeping the transformer parameters frozen and only learning the embedding and un-embedding parameters with a high learning rate ($10^{-3}$). We notice that this usually aids in reaching the same loss faster and can lead to small gains in the final performance. 

\paragraph{Continued Pre-Training}
We perform continued pre-training on each target language with 200K examples ($\sim$200M tokens) and 500K examples ($\sim$500M tokens) from the mC4 corpus \citep{raffel2023exploring} for larger (>6B parameters) and smaller models, respectively. The effective vocabulary sizes, and data statistics (in bytes) for the tokenizer training and continued pre-training are summarized in~\autoref{table:stats} in the appendix. For main experiments, we add 50K tokens when training on a larger CPT corpus (200K-500K documents). For other analyses, we use 10K tokens and train on 100K documents due to computational constraints.

\paragraph{Implementation Details} We train on 4 A40 GPUs for a single epoch with a cosine-warmup scheduler, max sequence length $1024$, with batch sizes that maximize memory usage, as presented in \autoref{tab:hparams} in the Appendix. For 200K examples, this translates to roughly 18h of training on all 4 GPUs. We use full-finetuning in all of our experiments, since training with LoRA \citep{hu2022lora} yielded much worse performance, and only led to $1.5\times$ less compute.

\section{Evaluation Setting}

\subsection{Experimental Goals}

\paragraph{Language Models}
\label{subsec:comparison_systems}

Table \ref{table:comparison_systems} summarizes the open-access base LLMs that we consider. XGLM-7.5B~\citep{lin2021few}, Bloom-7.1B~\citep{workshop2023bloom}, and Gemma models~\citep{gemmateam2024gemma} are equipped with a large vocabulary (251K-256K) that encompasses multiple languages.  XGLM was trained on CC100-XL \citep{lin2021few} on 500B tokens across 30 languages, Bloom was trained on ROOTS~\citep{laurençon2023bigscience} on 350B tokens across $\sim$46 languages, while Gemma-7B and LLaMA-2-7B were trained on 6T and 2T tokens, respectively. The vocabulary of Mistral-7B and LLaMA-2 are primarily monolingual, with about 32K tokens. The smaller models, Gemma-2B and Phi-2~\citep{microsoftPhi2Surprising}, were trained on 1.6T and 2T tokens, respectively. Gemma-7B variant far outperforms any existing open-sourced 7B LLMs in our evaluations on multilingual benchmarks.

\paragraph{Target Languages} We evaluate on four languages -- Hindi, Arabic, Turkish, and Tamil, which covers languages in Latin (Turkish) and non-Latin scripts (Hindi, Arabic, Tamil). Based on the performance of the base language models on downstream benchmarks, we group the languages into mid (Hindi, Arabic, Turkish) and low-resource (Tamil).\looseness=-1

\subsection{Evaluation Tasks}

We select a range of multilingual benchmark datasets to cover diverse tasks. Except for machine translation and headline generation, all other benchmarks are formatted as multiple-choice tasks to facilitate evaluation.~\autoref{table:appdx_eval_stats} in the Appendix reports dataset statistics.

 \begin{figure*}[t]
    \centering
    \begin{subfigure}{0.32\linewidth}
    \includegraphics[width=\linewidth]{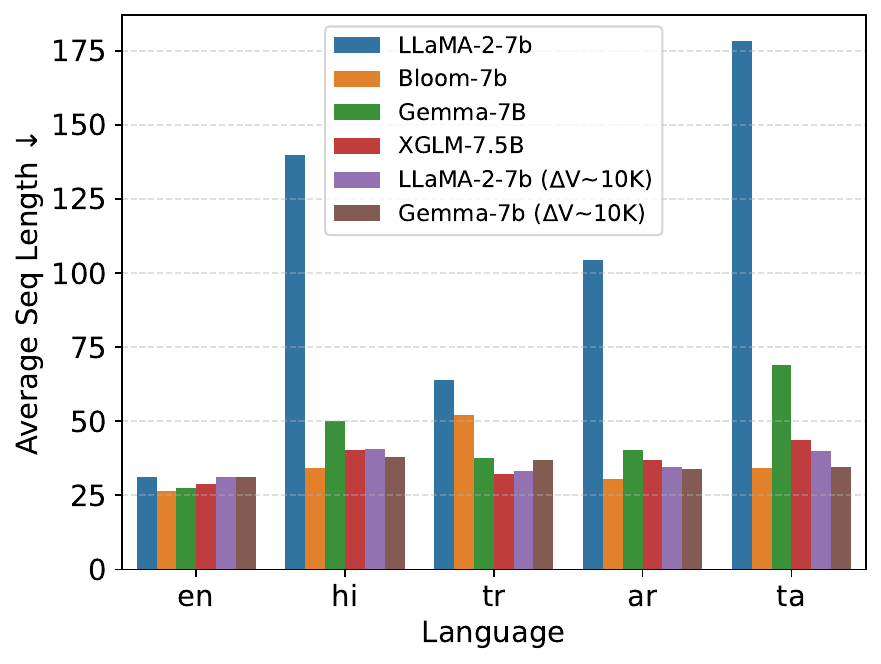}
    \caption{Fertility per Language \& LLM pair}
    \label{fig:lang_v_fertility}
    \end{subfigure}
    \begin{subfigure}{0.32\linewidth}
    \includegraphics[width=\linewidth]{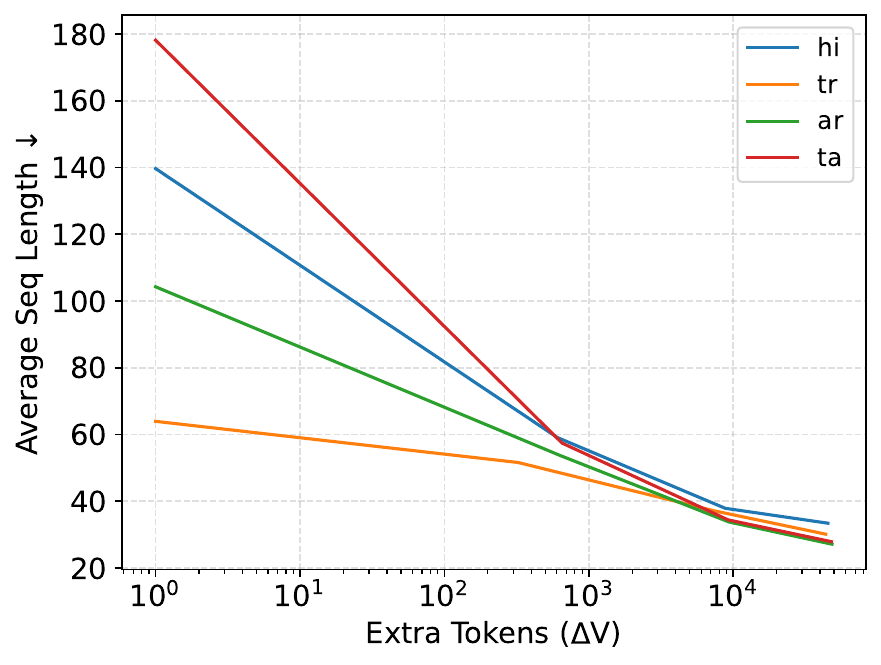}
    \caption{Fertility vs. Vocab Size}
    \label{fig:vocab_v_fertility}
    \end{subfigure}
    \begin{subfigure}{0.315\linewidth}
    \includegraphics[width=\linewidth]{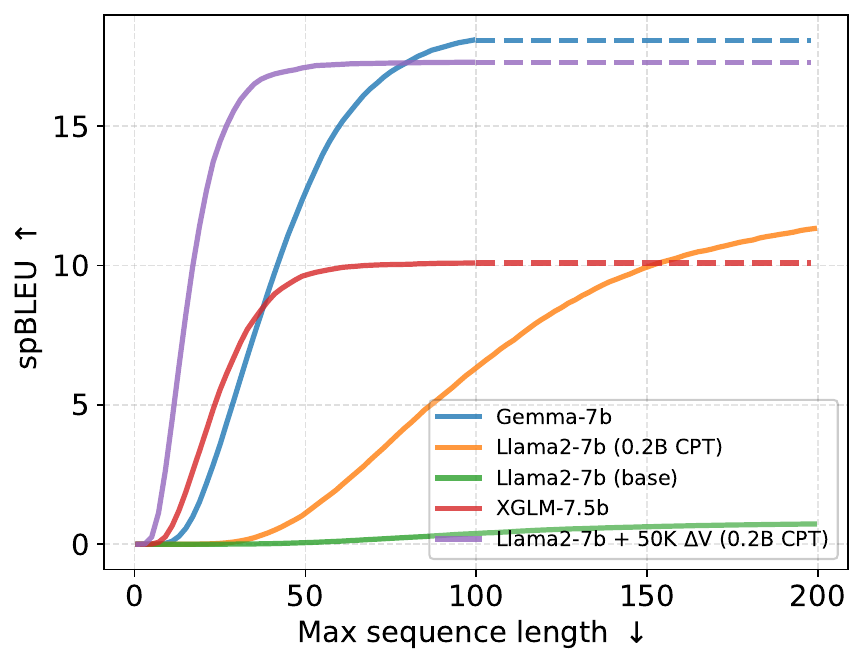}
    \caption{Performance vs. Fertility (Tamil)}
    \label{fig:bleu_vs_maxlen}
    \end{subfigure}
    \caption{Efficiency evaluation: the impact of vocabulary extension on the average sequence length. The shorter sequence length is more desirable. (a) extending the vocabulary with 10K tokens makes the token length substantially shorter, on par with that of English. (b) Adding more tokens continue to improve the sequence length with diminishing returns. (c) Performance (spBLEU) on FLORES machine translation versus maximum generation cutoff length; adapted models require 2$\times$ less tokens to achieve high performance.}
     \label{fig:impact_of_vocab_ext}
    \vskip -1ex
\end{figure*}

\paragraph{Generation Tasks}

\begin{itemize}[leftmargin=5mm,noitemsep]
    \item \textbf{Machine Translation}
We evaluate on the FLORES-200 benchmark \citep{nllbteam2022language}, which contains 1012 parallel examples in the test set. We perform 5-shot prompting with examples from the development set.
\item \textbf{Headline Generation}
We evaluate on XL-Sum \citep{xlsum}, which contains summaries and titles of news articles. We re-frame the task as generating titles from summaries, as the news articles are long. We prompt LLM with a single example from the training split. \\

\end{itemize}
\paragraph{Natural Language Understanding Tasks}

\begin{itemize}[leftmargin=5mm,noitemsep]
    \item \textbf{Knowledge Probing} We evaluate on mLAMA \citep{kassner-etal-2021-multilingual}, which contains knowledge triplets in the format $\langle$object, relation, subject$\rangle$. Following \citet{lin2021few}, we convert the triplets to a template that queries for the object, such as "Matthew Perry was born in [MASK]." Apart from the ground truth, we replace [MASK] with three other candidates that belong to the same relation.\looseness=-1
    \item \textbf{Natural Language Inference} We evaluate on XNLI \citep{conneau-etal-2018-xnli} in a 5-shot setting with the evaluation templates from prior work.\footnote{We use lm-evaluation-harness: \url{https://github.com/EleutherAI/lm-evaluation-harness}}
    \item \textbf{Sentiment Analysis} We use a recently released dataset \citep{doddapaneni-etal-2023-towards} of product reviews manually translated into Indic languages. The dataset contains parallel sentences in English and Indic languages. We generated an evaluation set for Turkish and Arabic through automated translation from English.\footnote{We use \url{https://github.com/ssut/py-googletrans}} We evaluate in a 5-shot setting.
    \item \textbf{Causal/Commonsense Reasoning} We use XCOPA \citep{ponti-etal-2020-xcopa} and XStoryCloze \cite{lin2021few}, using 5-shot in-context prompts for both tasks.
\end{itemize}

\subsection{Model Inference Setting}

We use the same prompt template for all models. For all multiple-choice datasets with training and evaluation splits (all datasets except for mLAMA), we select five input-output pairs from the respective training and evaluation splits to form in-context examples. The examples are selected randomly for each query to the LM, but with a fixed seed to maintain consistency across evaluations. For mLAMA, we use zero-shot prompts. The exact prompts, which include selected in-context exemplars, can be found in Appendix ~\ref{appendix:examples}.

For multiple-choice tasks, we select the continuation with the highest byte-length normalized log-probabilities, which is tokenizer agnostic \citep{eval-harness}. For generation tasks, we sample greedily, with a maximum of 200 tokens for translation and 50 tokens for summarization.

\subsection{Evaluation Metrics}

\paragraph{Efficiency Metrics} 
As we do not make changes to LLM architecture except for its tokenizer, our efficiency metric will focus on the number of tokens required to convey the same amount of information per language, following prior work~\citep{ahia2023all}. Specifically, we define the fertility as the average number of tokens required to encode a given text. This diverges from earlier work~\citep{rust-etal-2021-good} which defines fertility as the average number of sub-words per given word and make evaluation independent of word segmentation.   

For generation tasks, we additionally measure (\%Gen), the percentage of generated tokens that belong to the newly added vocabulary, $\Delta V$, and the number of examples processed per second (throughput). We define throughput as the number of examples processed by the model per second.

\paragraph{Task Performance Metrics} For the text-understanding benchmarks (multiple-choice questions), we measure the accuracy. For machine translation and summarization, we report spBLEU \citep{goyal-etal-2022-flores}, a universal version of BLEU that is comparable across languages.

\section{Results}
\label{sec:results}
We report performances on two axis -- efficiency (fertility) and end task performance.

\begin{table*}[t]
\small
\centering
\addtolength{\tabcolsep}{-3.7pt}
\begin{tabular}{llccccccc}
\toprule
\multirow{2}{*}{Lang}&\multirow{2}{*}{Base model} & \multicolumn{6}{c}{Task}\\
&  & \textbf{FLORES} & \textbf{XLSUM} & \textbf{MLAMA} & \textbf{Sentiment} & \textbf{XStoryCloze} & \textbf{XNLI} & \textbf{XCOPA} \\ \midrule \midrule
&Random Guess & - & - & {\small25.00} & {\small50.00} & {\small50.00} & {\small33.00} & {\small50.00} \\ \midrule 
\multirow{4}{*}{{\Large\textbf{\sfrac{hi}{tr}}}}& Gemma-7B & {\LARGE\sfrac{\textbf{32.26}}{\textbf{26.64}}} & {\LARGE\sfrac{\textbf{14.60}}{\underline{15.21}}} &{\LARGE\sfrac{\textbf{50.05}}{\underline{56.84}}} & {\LARGE\sfrac{\textbf{95.20}}{\textbf{97.20}}} & {\LARGE\sfrac{\textbf{70.62}}{-}} & {\LARGE\sfrac{\textbf{41.65}}{\underline{42.25}}} & {\LARGE\sfrac{-}{\textbf{69.00}}} \\
& Bloom-7.1B & {\LARGE\sfrac{21.32}{1.23}} & {\LARGE\sfrac{7.91}{7.87}} & {\LARGE\sfrac{\underline{47.21}}{40.52}} & {\LARGE\sfrac{\underline{94.00}}{64.70}} & {\LARGE\sfrac{64.99}{-}} & {\LARGE\sfrac{\textbf{41.65}}{34.02}} & {\LARGE\sfrac{-}{51.60}} \\
& XGLM-7.5B  & {\LARGE\sfrac{18.12}{16.75}} & {\LARGE\sfrac{8.66}{6.50}} & {\LARGE\sfrac{43.21}{55.96}} & {\LARGE\sfrac{82.40}{66.90}} & {\LARGE\sfrac{61.22}{-}} & {\LARGE\sfrac{\underline{40.72}}{\underline{41.49}}} & {\LARGE\sfrac{-}{60.00}}\\
\rowcol & LLaMA-2 {\tiny($\Delta V$=50K, CPT)} & {\LARGE\sfrac{\underline{28.15}}{\underline{20.95}}} & {\LARGE\sfrac{\underline{13.70}}{\textbf{16.03}}} & {\LARGE\sfrac{44.45}{\textbf{61.72}}}  &  {\LARGE\sfrac{87.60}{\underline{85.10}}} & {\LARGE\sfrac{\underline{67.90}}{-}} & {\LARGE\sfrac{39.36}{40.48}} & {\LARGE\sfrac{-}{\underline{61.80}}} \\ \midrule
\multirow{4}{*}{{\Large\textbf{\sfrac{ta}{ar}}}}& Gemma-7B & {\LARGE\sfrac{\textbf{18.08}}{\textbf{26.99}}} &
{\LARGE\sfrac{\underline{14.51}}{\underline{17.59}}} & {\LARGE\sfrac{\textbf{44.65}}{\underline{52.67}}} & {\LARGE\sfrac{\textbf{96.70}}{\textbf{97.60}}} & {\LARGE\sfrac{-}{\textbf{68.83}}} & {\LARGE\sfrac{-}{\textbf{39.60}}} & {\LARGE\sfrac{\underline{56.20}}{-}} \\
& Bloom-7.1B & {\LARGE\sfrac{8.74}{20.40}} & {\LARGE\sfrac{10.91}{13.51}} &{\LARGE\sfrac{41.41}{\textbf{53.55}}} & {\LARGE\sfrac{64.70}{85.40}} & {\LARGE\sfrac{-}{62.67}} & {\LARGE\sfrac{-}{\underline{37.51}}} & {\LARGE\sfrac{54.80}{-}} \\
& XGLM-7.5B & {\LARGE\sfrac{10.17}{12.23}} & {\LARGE\sfrac{3.86}{6.84}} &{\LARGE\sfrac{37.19}{47.18}} & {\LARGE\sfrac{76.10}{71.10}} & {\LARGE\sfrac{-}{58.37}} & {\LARGE\sfrac{-}{36.18}} & {\LARGE\sfrac{51.60}{-}} \\
\rowcol & LLaMA-2 {\tiny($\Delta V$=50K, CPT)} & {\LARGE\sfrac{\underline{17.29}}{\underline{22.33}}} & {\LARGE\sfrac{\textbf{16.37}}{\textbf{17.66}}} & {\LARGE\sfrac{\underline{41.55}}{47.00}} & {\LARGE\sfrac{\underline{93.20}}{\underline{89.70}}} & {\LARGE\sfrac{-}{\underline{64.59}}} & {\LARGE\sfrac{-}{35.82}} & {\LARGE\sfrac{\textbf{58.40}}{-}}\\
\bottomrule
\end{tabular}
\caption{Comparing adapted monolinugal model (LLaMA-2-7B) against state-of-the-art multilingual models. We report spBLEU for FLORES dataset and accuracy for all other datasets. \textbf{Bold} values indicate the best performance in for each dataset while \underline{underlined} values indicate the second best number. In this table, $\Delta V = \sim50K$ (i.e. 50K added tokens) and CPT done for 200M tokens. Our adapted models will have a separate checkpoint per language, while others use the same checkpoint for all languages.}
\label{table:perf_cmp}
\end{table*}

\begin{table*}[t]
\centering
\small
\addtolength{\tabcolsep}{-3.5pt}
\begin{tabular}{lcccccc}
\toprule
& \multicolumn{3}{c}{\Large\textbf{\sfrac{en $\rightarrow$ hi}{en $\rightarrow$ tr}}} & \multicolumn{3}{c}{\Large\textbf{\sfrac{en $\rightarrow$ ar}{en $\rightarrow$ ta}}} \\ \midrule
\textbf{Model} & \textbf{Throughput $\uparrow$} & \multicolumn{1}{r}{\textbf{Fertility}$\downarrow$} & \textbf{\%Gen} & \textbf{Throughput $\uparrow$} & \multicolumn{1}{r}{\textbf{Fertility}$\downarrow$} & \textbf{\%Gen} \\ \midrule \midrule
XGLM  &  {\LARGE\sfrac{0.71}{0.82}} & {\LARGE\sfrac{39.85}{\textbf{34.81}}} & - &  {\LARGE\sfrac{0.71}{0.57}} &  {\LARGE\sfrac{39.98}{48.84}} & - \\
Gemma-7B &  {\LARGE\sfrac{0.47}{0.63}} & {\LARGE\sfrac{52.34}{40.04}}  & - &  {\LARGE\sfrac{0.63}{0.32}} &  {\LARGE\sfrac{38.06}{77.14}} & - \\
LLaMA-2 + CPT  & {\LARGE\sfrac{0.18}{0.31}} &  {\LARGE\sfrac{154.52}{93.02}} & - & {\LARGE\sfrac{0.24}{0.14}} &  {\LARGE\sfrac{116.71}{192.13}} & -  \\
\rowcol LLaMA-2 {\tiny($\Delta V$=50K, CPT)}  & {\LARGE\sfrac{\textbf{0.85}}{\textbf{0.84}}} & {\LARGE\sfrac{\textbf{37.49}}{39.04}} &  {\LARGE\sfrac{81.72}{62.96}} & {\LARGE\sfrac{\textbf{1.06}}{\textbf{0.89}}} & {\LARGE\sfrac{\textbf{27.59}}{\textbf{32.87}}} & {\LARGE\sfrac{79.85}{70.43}}  \\ \midrule
\end{tabular}
\caption{Efficiency comparisons on FLORES machine translation task. \% Gen indicates the \% of tokens generated that belong to the extra vocabulary $\Delta V$. Fertility is the average sequence length of generations. Adapted LLaMA-2 achieves throughput that is up to 68\% higher than multilingual models.}
\label{table:efficiency}
\end{table*}

\subsection{Efficiency Analysis} 

\paragraph{Vocabulary augmentation effectively improves fertility in low-resource languages} We compare the fertility between tokenizers by measuring the average sequence length on target sentences from the FLORES benchmark. \autoref{fig:lang_v_fertility} shows a significant disparity in fertility between English and other languages, especially low-resource, before vocabulary adaptation. The disparity is more pronounced on primarily monolingual LMs. Similar disparity has been observed in commercial language models with multilingual capabilities~\citep{ahia2023all}. Augmenting the base vocabulary of these models with {\small $\sim$}10K language-specific tokens significantly mitigates the disparity, providing fertility of the target language that is on-par with that of English. For low-resource language (e.g., Tamil), even multilingual model (Gemma-7B) shows significant gain in fertility after vocabulary extension.\looseness=-1

\paragraph{Relationship between the vocab size and fertility} \autoref{fig:vocab_v_fertility} reports the relationship between the number of added tokens $|\Delta V|$ and fertility when extending LLaMA-2's vocabulary. We observe that fertility on the target-language improves with increasing vocabulary size, but with diminishing gains as the extra token increases. Adding {\small $\sim$}1K language-specific tokens doubles the fertility on average. The gains are more pronounced for non-latin scripts (hi, ar, ta), with up to $3$x increase after adding $\sim$1K tokens.\looseness=-1

\subsection{End Task Performance}  
\label{subsec:end_task_performance}
We have found that fertility can be improved with relatively modest amount of vocabularly extension. But can LLM make use of newly added vocabulary tokens efficiently, given they were unseen during its pre-training? To answer this question, we discuss end task performances. 
\paragraph{Adapted English models can match the performance of base multilingual models} Multilingual LLMs often show stronger performance on a wide range of language compared to primarily English LLMs. However, state-of-the-art English models are released more frequently with various configurations (model sizes, architecture, focus domains) than their multilingual equivalents. We study whether adapting these predominantly English models through vocabulary extension and continued pre-training (CPT) can enable them to match the performance of multilingual models. In \autoref{table:perf_cmp}, we compare the adapted LLaMA-2-7B model against XGLM-7.5B, Gemma-7B, and Bloom-7.1B. Notably, the recently released Gemma-7B model demonstrates superior performance in all tasks across four languages. 

LLaMA model performance improves substantially through pre-training on a relatively small target-language corpus. Our adapted LLaMA model, LLaMA-2-7B ($\Delta V$=50K, CPT), exhibits highly competitive performance with respect to Gemma 7B, particularly on the low-resource language (Tamil), with 1.5 billion fewer parameters. As shown in \autoref{fig:bleu_vs_maxlen}, LLaMA-2-7B ($\Delta V$=50K, CPT) competes with Gemma-7B using 2$\times$ less tokens.  Additionally, as illustrated in \autoref{table:efficiency}, the adapted model achieves up to 8 times higher throughput compared to LLaMA without vocabulary extension and also substantially higher than Gemma 7B. 
These enhancements are more pronounced for non-Latin and low-resource languages (Hindi, Arabic, Tamil) compared to Latin languages (Turkish). Moreover, adapted monolingual models produce up to 1.5 times$\times$ smaller sequences than XGLM and Gemma models.\looseness=-1

\section{Analysis}
\begin{figure}
    \centering 
    \includegraphics[width=\linewidth]{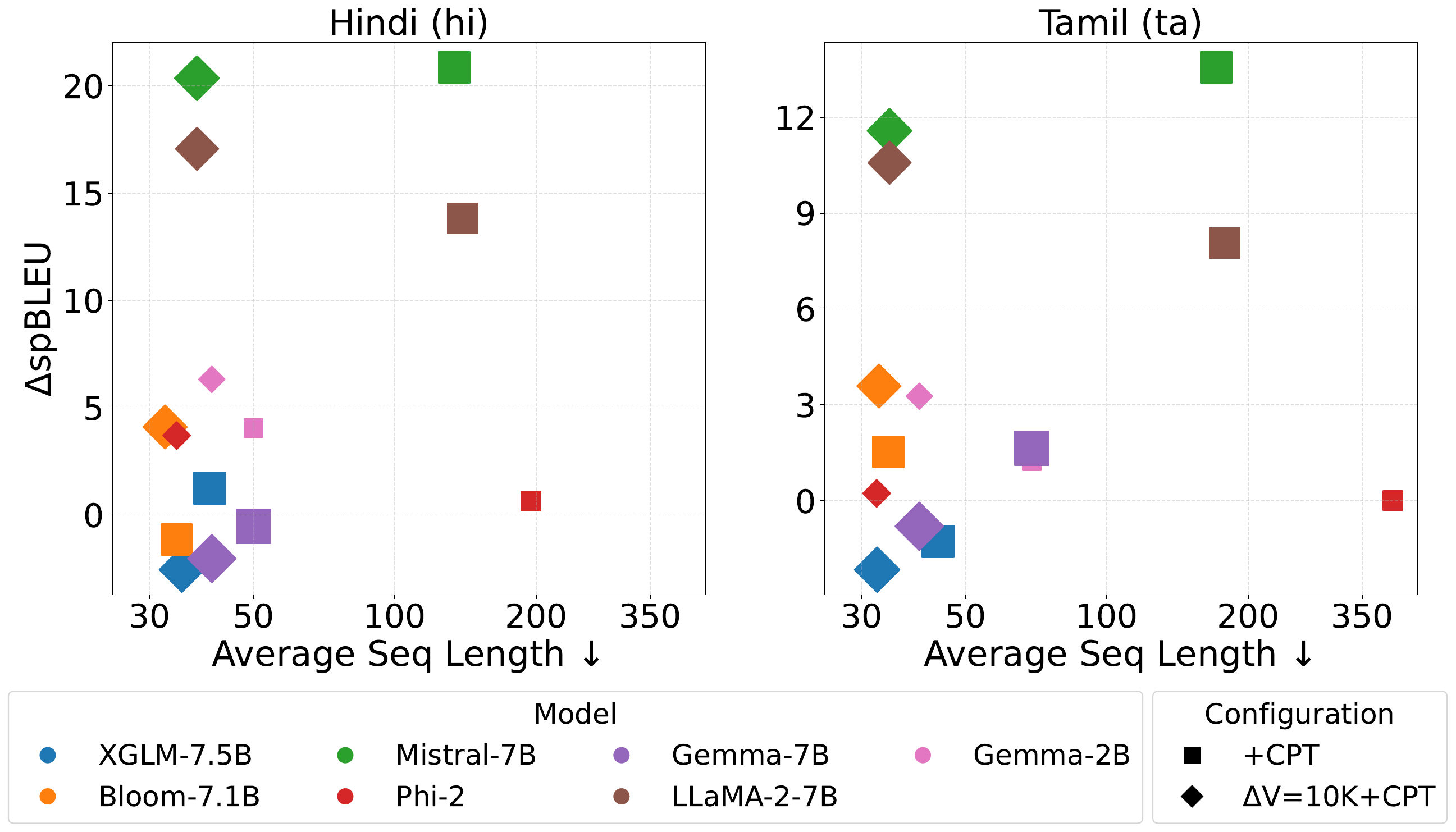}
    \caption{Change in performance ($\Delta$spBLEU) after adaptation. We report average performance across all benchmarks for two language pairs {\sfrac{hi}{ta}}, with continued training on 100K examples and vocabulary extension of $\Delta V$=10K tokens. Larger models are represented with bigger markers. Absolute numbers on all benchmarks are provided in \autoref{table:model_choice_hi} and \autoref{table:model_choice_ta} in \autoref{appdx:additional_exp}.}
     \label{fig:fertility_vs_perf}
\end{figure}

\paragraph{Can multilingual models benefit from the same adaptation recipe?} \autoref{fig:fertility_vs_perf} illustrates the change in performance ($\Delta$spBLEU) on the FLORES generation task observed when adapting various monolingual and multilingual models for Hindi (hi) and Tamil (ta). Here, we fix the adaptation recipe: we train on 100K examples, both with and without vocabulary augmentation ($\sim$10K tokens). We compare four kinds of base models - large monolingual (Mistral-7B, LLaMA-2-7B), small monolingual (Phi-2), large multilingual (XGLM-7.5B, Bloom-7.1B, Gemma-7B), and small multilingual (Gemma-2B).\looseness=-1

\paragraph{The lack of end task performance gains: Phi-2 and Gemma-7B} Larger monolingual models (LLaMA and Mistral) improve up to 12$\times$ over their base variants. Here, the more capable English model (Mistral) performs better after adaptation for both mid and low resource settings. The smaller, yet highly competitive, monolingual model i.e. Phi-2~\citep{microsoftPhi2Surprising, abdin2024phi3} exhibits minimal improvement, particularly on Tamil. Phi-2 is the only model in our comparisons that was trained on a curated, high quality monolingual data. This suggests that adapting monolingual model trained on a highly curated data might be more challenging.\looseness=-1

Smaller multilingual models like Gemma-2B still exhibit up to 56\% relative improvement, and show competitive end-task performance after adaptation. Gemma-7B, which was trained on extensive amounts of data (6T tokens), does not show notable performance enhancements from either continued pretraining (CPT) or vocabulary extension. This may indicate that its extensive pretraining saturates multilingual capabilities, rendering additional improvements through CPT or vocabulary expansion redundant. Similarly, XGLM shows limited benefit from these adaptations, likely due to its pre-training dataset that mirrors our CPT corpus -- suggesting that when the training data is already aligned with the adaptation data, further modifications may not yield substantial performance gains. In contrast, BLOOM, which is trained on a curated corpus of 46 languages~\citep{laurençon2023bigscience}, demonstrates relative improvements of up to 40\%.

\begin{figure}
     \centering
     \includegraphics[width=\linewidth]{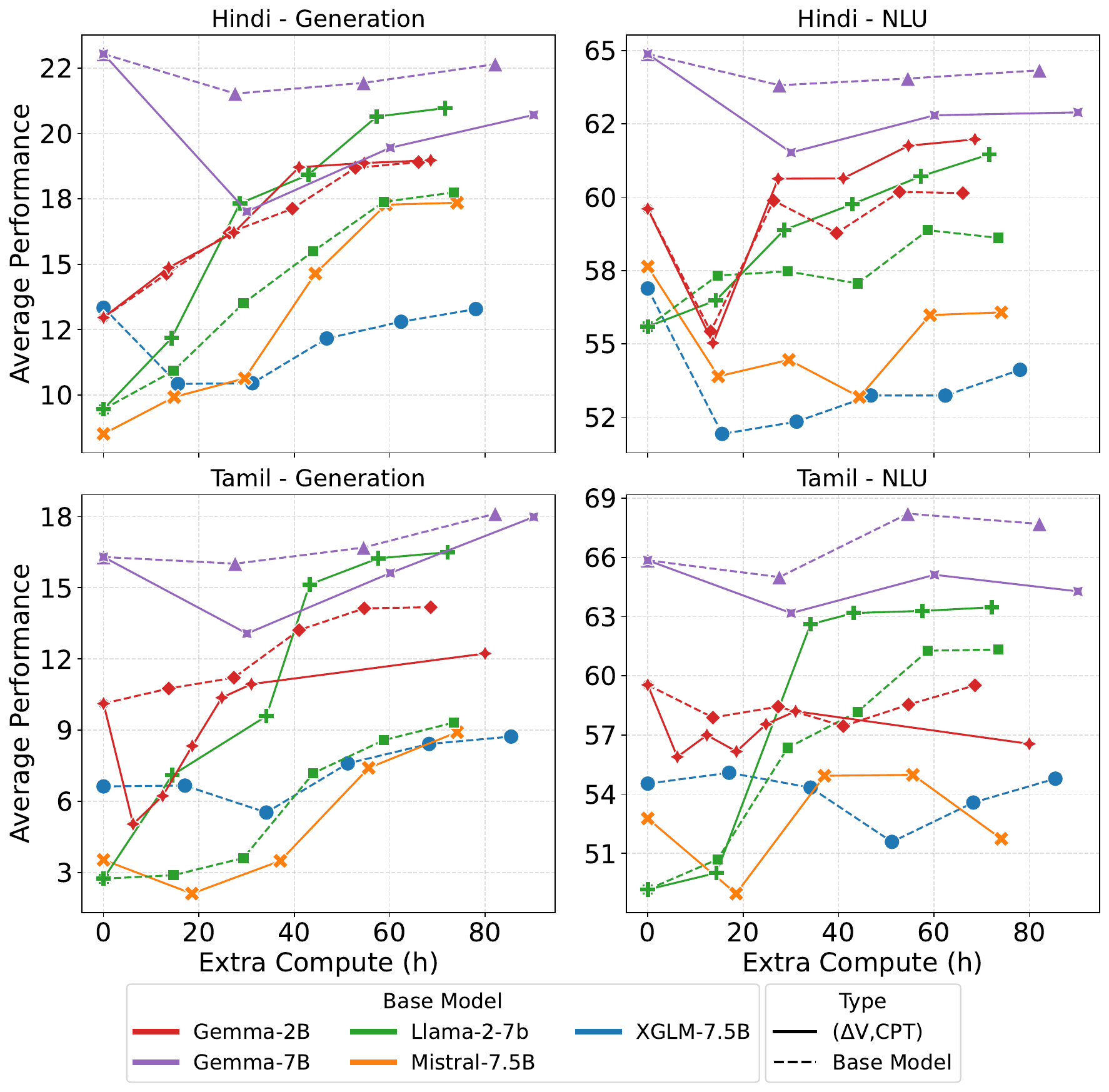}
     \caption{Adapted LLM's performance on Hindi/Tamil on generation and understanding benchmarks with increasing compute (measured in terms of hours per GPU). For models with extended vocabulary, $\Delta V$=50K.}
     \label{fig:perfvscompute_2}
\end{figure}

\begin{table*}[t]
\small
\centering
\begin{tabular}{@{}lcccccccc@{}}
\toprule
{\small\textbf{CPT data}} & {\small\textbf{V'}} & {\small\textbf{\#params}} & {\small\textbf{FLORES}} & {\small\textbf{XLSUM}} & {\small\textbf{MLAMA}} & {\small\textbf{Sentiment}} & {\small\textbf{XStoryCloze}} & {\small\textbf{XNLI}} \\ \midrule \midrule
\multirow{3}{*}{100K} & 1K & 6.7B & 23.46 & 8.65 & 43.44 & 90.20 & \textbf{67.24} & 40.00 \\
 & 10K & 6.8B & \textbf{25.24} & 11.99 & \textbf{44.14} & \textbf{92.10} & 65.65 & 41.04 \\
 & 50K & 7.1B & 25.02 & \textbf{12.36} & 42.45 & 91.50 & 67.11 &  \textbf{41.61}\\ 
 \midrule
\multirow{3}{*}{200K} & 1K & 6.7B & 26.50 & 9.23 & 45.08 & 91.20  & 65.65 & 39.40  \\ 
 & 10K & 6.8B &  27.31 & \textbf{12.45} & \textbf{45.62} & 86.60 & 65.12 & 40.48 \\
 & 50K & 7.1B & \textbf{27.86} & 12.36 & 43.87 & \textbf{94.10} & \textbf{67.84} & \textbf{41.41}\\ 
 \bottomrule
\end{tabular}
\caption{Performance on Hindi (LLama-2-7B) with increasing vocabulary size and CPT data (\# examples). With larger amount of data (200K), larger vocab size (50K) leads to performance gain, while with smaller amount of data (100K), smaller vocab (10K) often leads better performance.  \looseness=-1 }
\label{Table:vocab_size}
\end{table*}

\begin{table*}[t]
\small
\centering
\addtolength{\tabcolsep}{1pt}
\begin{tabular}{@{}lccccccc@{}}
\toprule
{\small\textbf{Model}} & {\small\textbf{FLORES}} & {\small\textbf{XLSUM}} & {\small\textbf{MLAMA}} & {\small\textbf{Sentiment}} & {\small\textbf{XStoryCloze}} & {\small\textbf{XNLI}}  \\ \midrule \midrule
\texttt{LLaMA-2-7B (base)} & 8.17 & 10.74 & 38.92 & 91.60 & 57.18  & 36.63 & \\ \midrule
\texttt{Random-Init} & 17.82 & 10.59 & 40.37 & 91.90 & 63.34  & 39.88 &  \\
\texttt{Random-Tok-Emb} & 17.01 & 9.22 & 39.09 & 78.90 & 62.67 & 39.68 &   \\
\texttt{Mean} & \textbf{25.24} & 11.99 & \underline{44.14}  & \underline{92.10} & 65.65 & \textbf{41.04} \\
\texttt{FOCUS} & 24.33 & \textbf{12.79} & 43.40 & \textbf{93.20} & \underline{66.78} & 38.59 \\
\texttt{Learned-Emb} & \underline{24.59} & \underline{12.41} & \textbf{44.67} & 86.50 & \textbf{67.50} & \underline{39.92}\\
 \bottomrule
\end{tabular}
\caption{Results on Hindi benchmarks (LLaMA-2-7B) with varying embedding initialization strategies ($\Delta V$=10K). Simple mean of constituent tokens performs competitively with respect to more complex methods.\looseness=-1}
\label{table:init_strategy}
\end{table*}

\paragraph{Training a smaller model with more data vs. a larger model with less data.} As shown in \autoref{fig:perfvscompute_2}, this choice depends heavily on the target language, compute budget, and the type of base model. For shorter training periods (<30h), the smaller multilingual model (Gemma-2B) is highly effective for both mid-resource (Hindi) and low-resource (Tamil) languages, making it a viable choice in scenarios where computational resources are limiting factors. While Gemma-7B achieves the highest performance across all tasks and languages, it consumes more computational overhead due to its larger size (up to 1B more than any other model we consider), making it less practical for resource-limited settings. 

With longer training times (>40h), LLaMA-2-7B surpasses Gemma-2B on generation tasks for Hindi (\autoref{fig:perfvscompute_2}), and across all tasks for Tamil. A larger target vocabulary size of 50K seems to hinder performance in the stronger English model (Mistral) as opposed to a smaller vocabulary size of 10K (\autoref{fig:fertility_vs_perf}). LLaMA-2-7B offers a good balance between adaptation computational efficiency and performance. This makes LLaMA-2-7B suitable for low-resource languages, particularly on generation tasks, when longer training periods are possible, but with a trade-off of increased inference time.\looseness=-1

\paragraph{Size of augmented vocabulary can be scaled proportionally to CPT data}
We compare performance of our model variants with increasing target vocabulary size in \autoref{Table:vocab_size}. In terms of efficiency, the bigger the vocabulary size, the fewer the number of tokens that are needed to encode the same amount of information. But how would it impact end task performance? We observe a more mixed result here: when the models are trained on smaller amounts of data (100K examples), there is no significant gain from adding more than 10K tokens to the base vocabulary. However, on doubling the training data, performance continues to grow with vocabulary size. Our findings align with \citet{dagan2024getting}, which studied vocabulary extension to adapt LLM to a code domain, that larger vocabulary sizes do not decrease downstream task performance when fine-tuning on billions of tokens. In our experiments, we observe that additional training only in the order of million tokens is sufficient for adaptation of monolingual models.

\paragraph{Mean embedding initialization is simple and effective}

We present the results of five different initialization strategies in \autoref{table:init_strategy}. \texttt{Random-Init} initializes the new token embeddings with random values, while \texttt{Random-Tok-Emb} uses the embedding of a randomly selected existing token for initialization. In the \texttt{Learned-Emb} strategy, we freeze all the model layers except for the embedding layer and train on 10\% of the CPT data before switching to full fine-tuning. \texttt{FOCUS}~\citep{dobler-de-melo-2023-focus} computes embeddings based on semantic similarity within an auxiliary static token embedding space. We observe that \texttt{FOCUS}, \texttt{Mean}, and \texttt{Learned-Emb} outperform the random initialization techniques. Notably, the simplicity of \texttt{Mean}, combined with its competitive efficacy, renders it an appealing choice for initialization. \texttt{Learned-Emb} suffers a performance drop on sentiment analysis, where the performance was high on this benchmark to begin with, and \texttt{Learned-Emb} might initialize embeddings that are farther away in the original embedding space as opposed to \texttt{Mean} and \texttt{FOCUS}. Our findings reinforce prior work showing that vocabulary initialization has big impact on downstream performances~\citep{minixhofer-etal-2022-wechsel, dobler-de-melo-2023-focus}.

\begin{table}[t]
\small
\centering
\addtolength{\tabcolsep}{-4.4pt}
\begin{tabular}{lcccc}
\toprule
\textbf{Model} & \textbf{MLAMA} & \textbf{Sent} & \textbf{XStory} & \textbf{XNLI} \\
\midrule \midrule
LLaMA-2-7B & 73.19 & 98.60 & 89.08 & 52.73 \\
LLaMA-2-7B {\tiny($\Delta V$=10K, CPT)} & 62.18 & 92.10 & 82.46 & 52.97 \\ \midrule
$\Delta$Perf. & -11.01 & -6.50 & 3.02 & 0.24 \\
\bottomrule
\end{tabular}
\caption{Evaluation of catastrophic forgetting on English benchmarks after continued pre-training on Hindi (hi) with 100K examples.}
\label{table:catastrophic_forgetting}
\end{table}

\paragraph{Does LLM lose performance in English after adaptation? Partially.} \autoref{table:catastrophic_forgetting} contrasts the downstream performance of our adapted model ($\Delta V$=10, CPT) with that of the base model (LLaMA-2-7B) on the source language (English). We observe that  performance on these tasks drops by 3.56\%, with the highest drop of 11\% on the knowledge probing tasks (MLAMA). Interestingly, we note that the cross-lingual task on the target language (English (en) $\rightarrow$ Hindi (hi)) is not significantly impacted by catastrophic forgetting, and in fact, improves after adaptation (\autoref{table:perf_cmp}).

\section{Conclusion}
In this paper, we presented a systematic study on the design choices involved in adapting LLMs to specific target languages. We explored the strategic choices in the adaptation process, such as the necessity of vocabulary extension, the addition of a large pool of language-specific tokens, and the criticality of initializing new parameters effectively. We contextualize these choices with respect to the end goal, such as the target language and a specified compute budget. Through comprehensive experiments on up to four languages and seven base language models, we showed that the efficiency of the model on a target language can be improved by a simple vocabulary augmentation step followed by further training. We further show that under the same adaptation strategy, adapted smaller multilingual models can be as good as their larger counterparts, and larger monolingual LMs can perform almost as good as multilingual LMs with relatively minimal amounts of target language data.

\subsection*{Limitations}
In this study, we are restricted to analyzing four languages due to computational limitations. In future research, we aim to explore a much wider range of languages, and a larger set of base models. We also did not explore trends on families of monolingual models of varying sizes. The vocabulary extension approach we study is limited to a straightforward union of vocabularies, and we did not investigate the factors in the BPE training corpus that might affect the generated tokens. We also did not investigate alternative tokenization approaches besides BPE. 

\subsubsection*{Acknowledgments}
We would like to thank Adam Klivans for providing compute resources. We also thank the members of UT NLP community and reviewers for feedback. This work was in part supported by Cisco Research. Any opinions, findings and conclusions, or recommendations expressed in this material are those of the authors and do not necessarily reflect the views of Cisco Research.

\bibliography{custom, references}

\clearpage
\appendix
\section*{Appendix}
The appendix is organized as follows:
\begin{itemize}
    \item In \autoref{appdx:train_setup}, we present the training setup and other details, such as CPT dataset statistics.\looseness=-1
    \item In \autoref{appdx:eval_settings}, we describe evaluation data statistics and evaluation setup such as prompt templates.
    \item In \autoref{appdx:additional_exp}, we present additional results and experiments.
\end{itemize}

\section{Training Setup}
\label{appdx:train_setup}

\begin{table}[h]
\small
\centering
\addtolength{\tabcolsep}{-2.5pt}
\begin{tabular}{@{}lccccc@{}}
\toprule
\textbf{Language} & \multicolumn{1}{c}{\textbf{\#Bytes: Tok}} & \textbf{|V'|} & \textbf{|V\textsubscript{new|}} & \textbf{\#Bytes: CPT}\\ \midrule
\multirow{3}{*}{Hindi (hi)} & \multirow{3}{*}{1.4B} & 1K & 32,613 & 0.5B &  \\
 &  & 10K & 40,816 & 0.8B &  \\
 &  & 50K & 77,338 & 0.9B &  \\ \midrule
\multirow{3}{*}{Turkish (tr)} & \multirow{3}{*}{0.9B} & 1K & 32,331 & 0.2B &  \\
 &  & 10K & 39,516 & 0.3B &  \\
 &  & 50K & 75,773 & 0.4B &  \\ \midrule
\multirow{3}{*}{Arabic (ar)} & \multirow{3}{*}{1.4B} & 1K & 32,638 & 0.4B &  \\
 &  & 10K & 41,337 & 0.6B &  \\
 &  & 50K & 80,348 & 0.7B &  \\ \midrule
\multirow{3}{*}{Tamil (ta)} & \multirow{3}{*}{2.6B} & 1K & 32,660 & 0.6B &  \\
 &  & 10K & 41,233 & 1.1B &  \\
 &  & 50K & 79,967 & 1.3B &  \\ \bottomrule
\end{tabular}
\caption{Our training dataset statistics. \#Bytes: Tok indicates the total number of bytes used to train the language-specific tokenizers, and \#Bytes: CPT indicate the effective number of bytes seen by the model during continued pre-training (CPT). $|V'|$ and $|V_{\text{new}}|$ indicate the vocab sizes of the language-specific vocabularies and merged vocabularies (\cref{subsec:augmenting_vocab}), respectively.}
\label{table:stats}
\end{table}

\begin{table}[h]
\small
\centering
\begin{tabular}{lccccccc}\toprule
Model & Learning Rate & Batch Size \\ \midrule
LLaMA-2-7B & 6E-05 & 8 \\ 
Gemma-7B & 3E-06 & 4 \\ 
Gemma-2B & 9E-05 & 16 \\ 
Mistral-7B & 3E-05 & 6 \\ 
Bloom-7.1B & 6E-05 & 8 \\ 
XGLM-7.5B & 3E-05 & 6 \\ 
Phi-2 & 9E-05 & 16 \\ 
\bottomrule
\end{tabular}
\caption{CPT hyperparameter setup for each model, along with per-device batch size. We manually tune hyperparameters. Due to compute limitations, we report numbers on single runs.}
\label{tab:hparams}
\end{table}

\subsection{FOCUS Implementation} We follow the codebase in \url{https://github.com/konstantinjdobler/focus}, which uses pre-trained fastText embeddings to initialize new embeddings.

\section{Evaluation Settings}
\label{appdx:eval_settings}

The statistics for each evaluation dataset are reported in \autoref{table:appdx_eval_stats}.
\begin{table}[h]
\small
\centering
\addtolength{\tabcolsep}{-2pt}
\begin{tabular}{@{}lcccccccc@{}}
\toprule
\textbf{} & \textbf{License} & \textbf{hi} & \textbf{tr} & \textbf{ar} & \textbf{ta} \\ \midrule 
\textbf{FLORES} & \tiny{CC-BY-SA-4.0} & 1012 & 1012 & 1012 & 1012 \\ 
\textbf{MLAMA} & \tiny{CC-BY-NC-SA-4.0} & 8570 & 14209 & 19354 & 7223 \\ 
\textbf{Sentiment} & \tiny{CC-0} & 1000 & 1000 & 1000 & 1000 \\ 
\textbf{XStoryCloze} & \tiny{CC-BY-SA-4.0} & 1511 & - & 1511 & - \\ 
\textbf{XNLI} & \tiny{CC-BY-NC-4.0} & 5010 & 5010 & 5010 & - \\ 
\textbf{XCOPA} & \tiny{CC-BY-4.0} & - & 500 & - & 500 \\ 
\textbf{XLSUM} & \tiny{CC-BY-NC-SA-4.0} & 8847 & 3397 & 4689 & 2027 \\ \bottomrule
\end{tabular}
\caption{Datasets: license information and statistics -- number of test examples for each benchmark \& language.}
\label{table:appdx_eval_stats}
\end{table}

\begin{figure}[t]
    \centering
     \includegraphics[width=\linewidth]{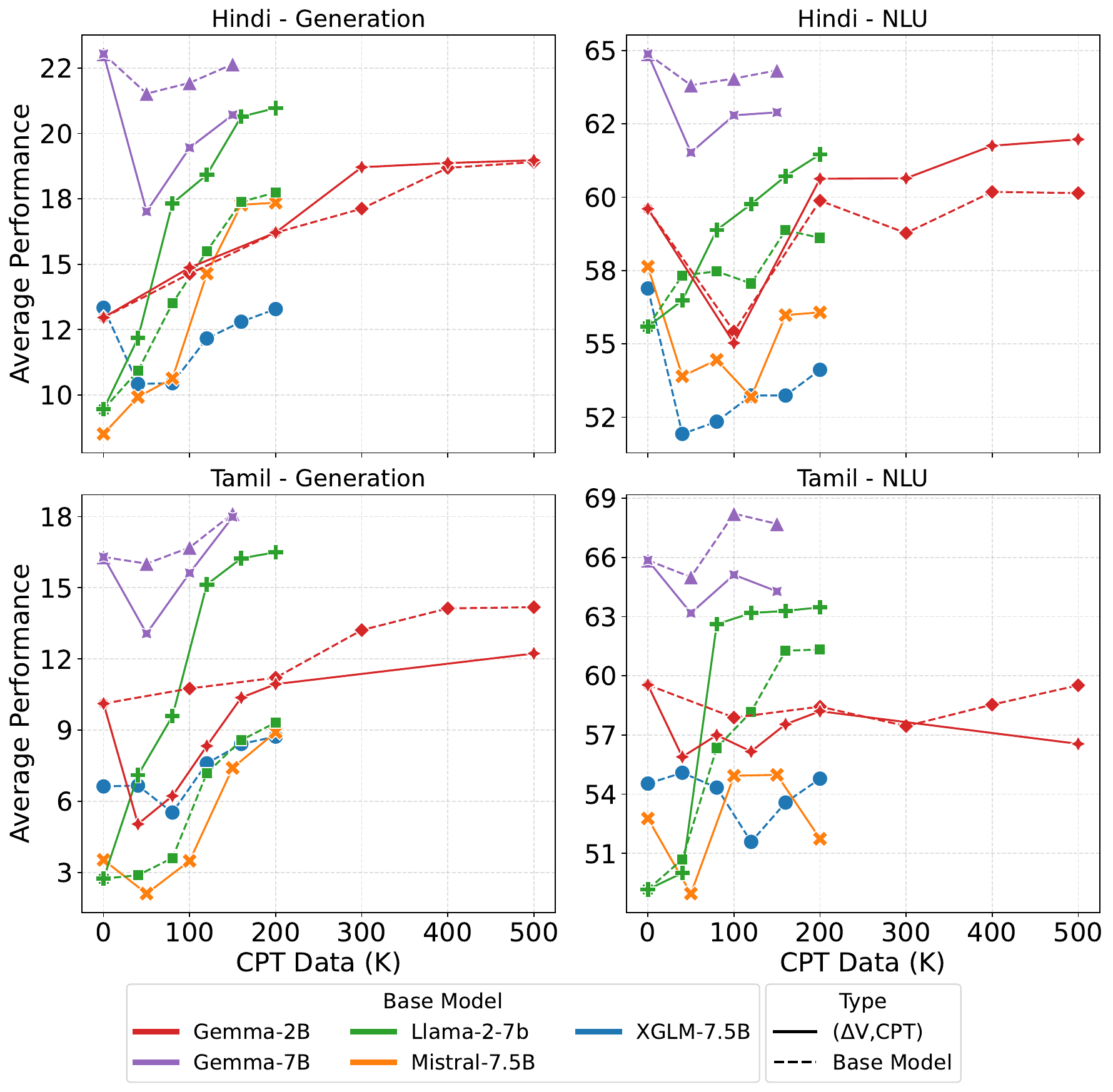}
    \caption{Performance variation on Hindi benchmarks with increasing CPT data (\#examples). For models with extended vocabulary (LLaMA, Mistral), $\Delta V$=50K.}
     \label{fig:perf_vs_data}
\end{figure}

\begin{table}[t]
\small
\addtolength{\tabcolsep}{-3.5pt}
\begin{tabular}{@{}lcccc@{}}
\toprule
\textbf{} & \textbf{Bloom-7.1B} & \textbf{Bloom-7.1B {\tiny(CPT)}} & \textbf{\begin{tabular}[c]{@{}c@{}}Bloom-7.1B\\ \tiny{$\Delta V$=50K, CPT}\end{tabular}} \\ \midrule
\textbf{Fertility} & 52.34 & 52.34 & 39.21 \\ \midrule
\textbf{FLORES} & 1.23 & 8.96 & 12.26 \\
\textbf{MLAMA} & 40.52 & 58.93 & 58.25 \\
\textbf{Sentiment} & 64.70 & 53.80 & 77.60 \\
\textbf{XNLI} & 34.02 & 40.08 & 37.79 \\
\textbf{XCOPA} & 51.60 & 55.40 & 54.00 \\ 
\rowcol \textbf{Average} & 40.74 & 44.92 & 46.52 \\ \bottomrule
\end{tabular}
\caption{Performance on Bloom-7.1B after adaptation on Turkish (unseen language). We train further on 100K examples.}
\label{table:bloom_tr_appdx}
\end{table}

\subsection{Prompt Templates}
\label{appendix:examples}
\begin{figure*}[]
     \centering
     \includegraphics[width=\textwidth]{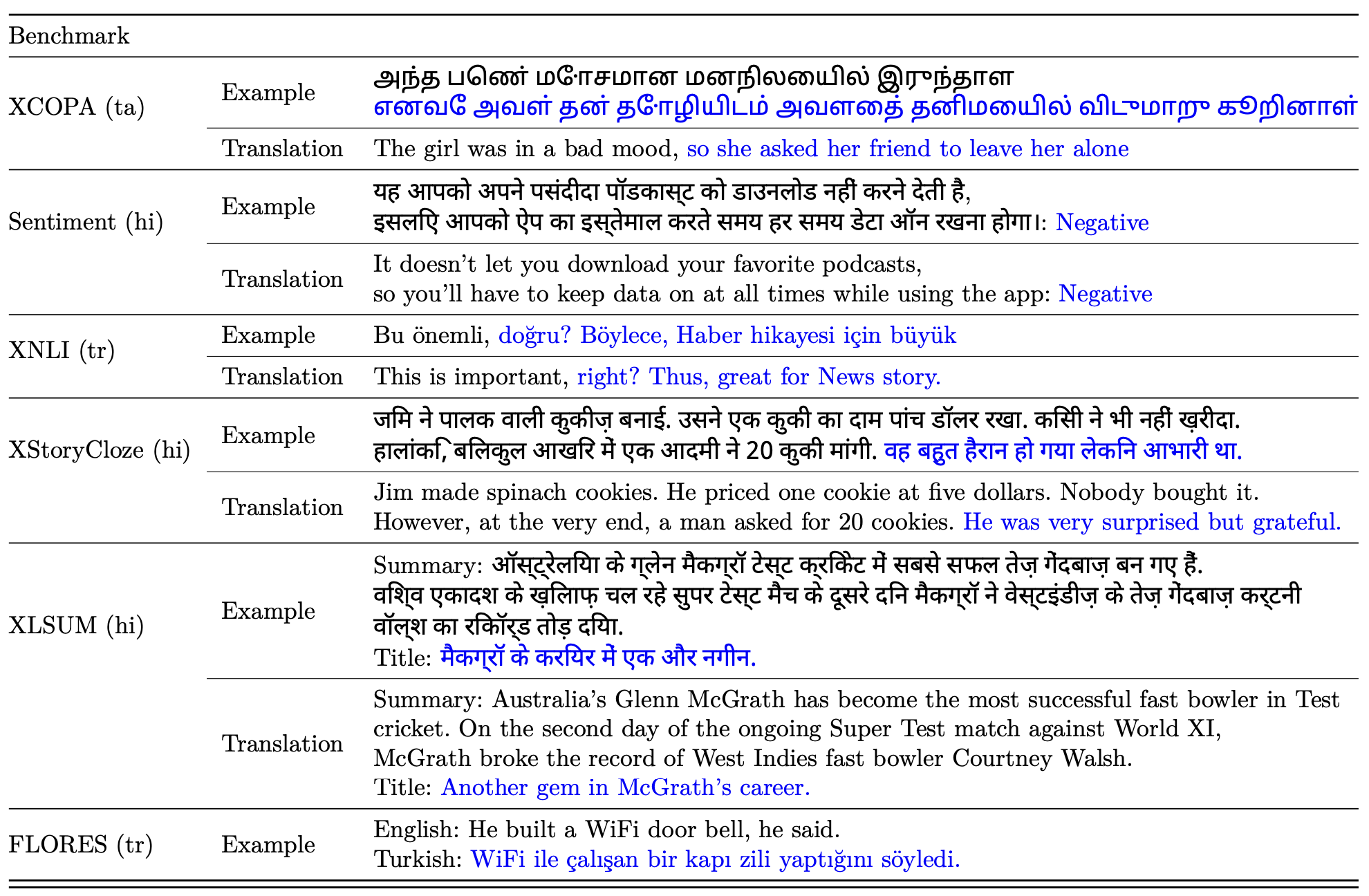}
     \caption{A single in-context example/template from each respective benchmark. \textcolor{blue}{Blue} indicates the continuation.}
     \label{fig:few_shot_examples}
\end{figure*}

For each task, we formulate the prompt as follows. Examples are provided in \autoref{fig:few_shot_examples}.

\noindent\textbf{Sentiment}: \textit{"sentence: [sentiment]"}, \textit{sentiment} $\in$ \{\textit{Positive, Negative}\} \\
\textbf{XNLI}: \textit{"premise. [connector]? hypothesis"}, \textit{connector} $\in$ \{\textit{Yes, No, Also}\} (in the target language) \\
\textbf{XStoryCloze}: \textit{"sentence, [continuation]"} \textit{continuation} $\in \mathcal{C}$, set of candidate continuations. \\
\textbf{XCOPA}: \textit{"premise. [continuation]"}, \textit{continuation} $\in \mathcal{C}$, set of candidate continuations. \\
\textbf{FLORES}: \textit{"English: [English Text], [Target Language]: [Translated Text]"}\\
\textbf{XLSUM}: \textit{"Summary: [Summary Text], Title: [Title Text]"}\\

\begin{table}[t]
\small
\centering
\addtolength{\tabcolsep}{-3.5pt}
\begin{tabular}{lccccc}
\toprule
Method & \textbf{FLORES} & \textbf{MLAMA} & \textbf{XCOPA} & \textbf{Sent} & \textbf{XLSUM} \\
\midrule
FOCUS & 10.61 & 38.06 & 56.20 & 93.10 & 16.76 \\
Mean & 11.32 & 42.12 & 55.00 & 93.10 & 16.61 \\
\bottomrule
\end{tabular}
\caption{Performance on Tamil Benchmarks (LLaMA-2-7B) with varying embedding initializaiton strategies ($\Delta V$=10K.}
\label{table:init_appdx}
\end{table}

\begin{table*}[t]
\small
\centering
\addtolength{\tabcolsep}{-3.2pt}
\begin{tabular}{@{}lccccccc@{}}
\toprule
\textbf{Model} & \textbf{FLORES} & \textbf{XLSUM} & \textbf{MLAMA} & \textbf{Sentiment} & \textbf{XStoryCloze} & \textbf{XNLI} & \textbf{Average} \\ \midrule
Phi-2      & 0.30  & 3.48   & 38.90 & 54.90 & 52.22 & 34.66 & 30.74 \\
Gemma-2B   & 15.52 & 10.40  & 44.47 & 92.20 & 62.14 & 44.47 & 44.87 \\
XGLM-7.5B  & 19.35 & 8.55   & 43.21 & 82.40 & 61.22 & 40.72 & 42.58 \\
Bloom-7.1B & 21.32 & 9.48   & 47.21 & 94.00 & 64.99 & 41.65 & 46.44 \\
Gemma-7B   & 32.26 & 14.60  & 50.05 & 97.20 & 70.62 & 41.65 & 51.06 \\
Mistral-7B & 5.35  & 11.67  & 42.50 & 92.50 & 60.36 & 41.02 & 42.23 \\
LLaMA-2-7B & 8.17  & 10.74  & 38.92 & 89.30 & 57.45 & 36.63 & 40.20 \\
Phi-2 {\tiny(CPT)} & 0.96  & 4.68   & 39.84 & 56.00 & 53.28 & 33.57 & 31.39 \\
Gemma-2B {\tiny(CPT)}  & 19.58 & 9.69   & 43.21 & 79.90 & 61.35 & 37.23 & 41.83 \\
XGLM-7.5B {\tiny(CPT)} & 20.61 & 6.64   & 43.86 & 69.40 & 59.56 & 40.44 & 40.09 \\
Bloom-7.1B {\tiny(CPT)} & 20.18 & 10.36  & 48.54 & 89.50 & 64.26 & 40.36 & 45.53 \\
Gemma-7B {\tiny(CPT)}  & 31.75 & 12.12 & 49.24 & 98.20 & 69.69 & 39.04 & 50.01 \\
Mistral-7B {\tiny(CPT)} & 26.21 & 14.06  & 47.41 & 96.20 & 69.29 & 40.56 & 48.96 \\
LLaMA-2-7B {\tiny(CPT)} & 21.99 & 10.90  & 45.02 & 92.40 & 64.06 & 39.36 & 45.62 \\
Phi-2 {\tiny($\Delta V$=10K, CPT)}     & 4.01  & 8.05   & 40.84 & 66.30 & 58.70 & 35.10 & 35.50 \\
Gemma-2B {\tiny($\Delta V$=10K, CPT)}  & 21.85 & 9.86   & 43.42 & 92.80 & 63.53 & 40.16 & 45.27 \\
XGLM-7.5B {\tiny($\Delta V$=10K, CPT)} & 16.81 & 5.63   & 43.79 & 70.90 & 61.02 & 40.64 & 39.80 \\
Bloom-7.1B {\tiny($\Delta V$=10K, CPT)} & 25.43 & 11.18  & 48.02 & 86.50 & 64.20 & 39.36 & 45.78 \\
Gemma-7B {\tiny($\Delta V$=10K, CPT)}  & 30.24 & 12.23 & 45.86 & 96.40 & 70.62 & 41.00 & 49.39 \\
Mistral-7B {\tiny($\Delta V$=10K, CPT)} & 25.21 & 12.14  & 43.42 & 81.90 & 65.32 & 41.97 & 44.99 \\
LLaMA-2-7B {\tiny($\Delta V$=10K, CPT)} & 25.24 & 11.99  & 44.14 & 92.10 & 65.65 & 41.04 & 46.69 \\ \bottomrule
\end{tabular}
\caption{Adapted LLM’s performance on Hindi, with continued fine-tuning on 100K examples.}
\label{table:model_choice_hi}
\end{table*}
\begin{table*}[t]
\small
\centering
\addtolength{\tabcolsep}{-3.2pt}
\begin{tabular}{lcccccc}
\toprule
\textbf{Model} & \textbf{FLORES} & \textbf{XLSUM} & \textbf{MLAMA} & \textbf{Sentiment} & \textbf{XCOPA} & \textbf{Average} \\ \midrule
Phi-2      & 0.02  & 0.46   & 34.43 & 49.30 & 48.20 & 26.48 \\
Gemma-2B   & 5.65  & 14.58  & 39.71 & 90.10 & 48.80 & 39.77 \\
XGLM-7.5B  & 10.17 & 4.36   & 37.19 & 76.10 & 51.60 & 35.88 \\
Bloom-7.1B & 8.47  & 11.38  & 41.41 & 64.70 & 54.80 & 36.15 \\
Gemma-7B   & 18.08 & 14.51  & 44.65 & 96.70 & 56.20 & 46.03 \\
Mistral-7B & 1.08  & 5.99   & 35.11 & 75.60 & 47.60 & 33.08 \\
LLaMA-2-7B & 1.54  & 3.95   & 36.81 & 62.10 & 48.60 & 30.60 \\
Phi-2 {\tiny(CPT)}     & 0.02  & 0.37   & 35.64 & 49.50 & 45.40 & 26.19 \\
Gemma-2B {\tiny(CPT)}  & 6.88  & 14.88  & 38.45 & 82.40 & 52.80 & 39.08 \\
XGLM-7.5B {\tiny(CPT)} & 8.90  & 7.11   & 39.71 & 61.60 & 53.00 & 34.06 \\
Bloom-7.1B {\tiny(CPT)} & 10.27 & 14.53  & 44.23 & 52.20 & 58.40 & 35.93 \\
Gemma-7B  {\tiny(CPT)} & 19.72 & 13.67 & 43.64 & 97.20 & 63.80 & 47.60 \\
Mistral-7B {\tiny(CPT)} & 14.64 & 9.39   & 38.49 & 89.60 & 58.80 & 42.18 \\
LLaMA-2-7B {\tiny(CPT)} & 8.81  & 7.41   & 38.72 & 82.00 & 53.60 & 38.11 \\
Phi-2 {\tiny($\Delta V$=10K, CPT)}    & 0.25  & 10.64  & 36.91 & 53.60 & 49.60 & 30.20 \\
Gemma-2B {\tiny($\Delta V$=10K, CPT)}  & 8.92  & 12.91  & 38.70 & 91.10 & 53.40 & 41.01 \\
XGLM-7.5B {\tiny($\Delta V$=10K, CPT)}  & 8.01  & 6.14   & 39.73 & 68.80 & 51.60 & 34.86 \\
Bloom-7.1B {\tiny($\Delta V$=10K, CPT)} & 12.33 & 14.69  & 44.57 & 51.00 & 57.20 & 35.96 \\
Gemma-7B {\tiny($\Delta V$=10K, CPT)}   & 17.28 & 16.28 & 42.34 & 97.10 & 58.20 & 46.24 \\
Mistral-7B {\tiny($\Delta V$=10K, CPT)} & 12.66 & 17.34  & 39.21 & 91.80 & 58.60 & 43.92 \\
LLaMA-2-7B {\tiny($\Delta V$=10K, CPT)} & 11.32 & 16.61  & 42.12 & 93.10 & 55.00 & 43.63 \\ \bottomrule
\end{tabular}
\caption{Adapted LLM’s performance on Tamil, with continued fine-tuning on 100K examples.}
\label{table:model_choice_ta}
\end{table*}

\section{Additional Experiments}
\label{appdx:additional_exp}

\paragraph{Impact of CPT and vocabulary extension on various multilingual and monolingual models} 
In \autoref{table:model_choice_hi} and \autoref{table:model_choice_ta} we present the full performance of all base models after adaptation -- without and with vocabulary augmentation. We observe results that are consistent with what we discuss in \autoref{subsec:end_task_performance}. Adapted primarily monlingual models (LLaMA, Mistral) show significant gains and match, if not surpass, most multilingual variants.

\paragraph{Given a fixed amount of data, train a smaller multilingual model or larger monolingual model?} As shown in \autoref{fig:perf_vs_data}, when given a fixed amount of data rather than a compute budget, training a larger model (either monolingual i.e LLaMA-2 or Gemma-7B) triumphs over tuning Gemma-2B, which needs more data. Again, the more capable English model (Mistral) does not do very well on a large amount of added tokens (50K).

\paragraph{Does Vocabulary Extension with CPT benefit multilingual models on unseen languages?} We report performance on Bloom-7.1B with vocabulary extension and further training in \autoref{table:bloom_tr_appdx}. As shown in \autoref{fig:lang_v_fertility}, the fertility on Turkish for Bloom is sub-optimal as it is an unseen language. We observe that vocabulary augmentation followed by CPT improves ferility by 25\%. On the generation tasks, we observe a relative performance improvement of 36\%, and 3\% improvement on all tasks on average as compared to just CPT.

\end{document}